\newcounter{mode}
\documentclass[10pt,twocolumn,letterpaper]{article}

\usepackage{cvpr}              
\usepackage{multirow}
\usepackage{multicol}
\usepackage{graphicx}
\usepackage{amsmath}
\usepackage{algorithmic}
\usepackage{amssymb}
\usepackage{booktabs}
\usepackage[numbers]{natbib}
\usepackage[ruled,vlined,linesnumbered]{algorithm2e}
\SetKwInput{KwIn}{Input}
\SetKwInput{KwPara}{Parameters}
\usepackage{overpic}
\usepackage{lipsum}
\usepackage{bbding}
\usepackage{pifont}
\usepackage[inline]{enumitem}

\usepackage[pagebackref,breaklinks,colorlinks]{hyperref}

\newcommand{\tbdline}{\textcolor{red}{TBD-TBD-TBD-TBD-TBD-TBD-TBD-TBD-TBD-TBD-TBD-TBD-TBD-TBD-TBD-TBD-TBD-TBD-TBD-TBD}\\}

\newcommand{\tbd}[1][1]{ \newcount\tmp \tmp=0 \loop \advance\tmp by 1 \tbdline \ifnum\tmp<#1 \repeat }

\newcommand{\figref}[1]{Fig.~\ref{#1}}
\newcommand{\tabref}[1]{Tab.~\ref{#1}}

\newcommand{\mathset}[1]{\mathcal{#1}}
\newcommand*\samethanks[1][\value{footnote}]{\footnotemark[#1]}

\usepackage[capitalize]{cleveref}
\crefname{section}{Sec.}{Secs.}
\Crefname{section}{Section}{Sections}
\Crefname{table}{Table}{Tables}
\crefname{table}{Tab.}{Tabs.}

\def\eg{\emph{e.g.}}

\def\cvprPaperID{5296} 
\def\confName{CVPR}
\def\confYear{2023}

\def\improvea#1{{\footnotesize (+#1)}}
\def\improveb#1{{\footnotesize  \color[rgb]{0.27, 0.71, 0.45} (+#1)}}
\def\textitf#1{{\footnotesize \textit{#1}}}
\usepackage[
skip=5pt]{caption}
\ifnum \value{mode}>1 
\usepackage{comment}
\onecolumn

\excludecomment{figure}

\excludecomment{figure*}
\expandafter\let\csname endfigure*\endcsname\relax

\excludecomment{table}

\excludecomment{table*}
\expandafter\let\csname endtable*\endcsname\relax

\excludecomment{equation}

\excludecomment{algorithm}

\providecommand{\figref}[1]{figure}
\renewcommand{\figref}[1]{figure}
\providecommand{\tabref}[1]{table}
\renewcommand{\tabref}[1]{table}
\providecommand{\equref}[1]{equation}
\renewcommand{\equref}[1]{equation}
\renewcommand{\ref}[1]{TBD}  

\renewcommand{\cite}[1]{\relax}
\providecommand{\bibliographystyle}[1]{\relax}
\renewcommand{\bibliographystyle}[1]{\relax}
\providecommand{\bibliography}[1]{\relax}
\renewcommand{\bibliography}[1]{\relax}

\else 

\fi

\begin{document}


\title{Curricular Object Manipulation in LiDAR-based Object Detection}

\author{Ziyue Zhu\textsuperscript{1}\thanks{The first two authors have equal contribution to this work.} \quad 
        Qiang Meng\textsuperscript{2}\samethanks \quad 
        Xiao Wang\textsuperscript{2} \quad 
        Ke Wang\textsuperscript{2} \quad
        Liujiang Yan\textsuperscript{2} \quad  
        Jian Yang\textsuperscript{1}\thanks{Jian Yang is the corresponding author (csjyang@nankai.edu.cn).}\\
        \textsuperscript{1}Tianjin Key Laboratory of Visual Computing and Intelligent Perception, \\
        College of Computer Science, Nankai University \quad
        \textsuperscript{2}Didi Chuxing\\
        {\tt\small
        zhuziyue@mail.nankai.edu.cn 
        \quad irvingmeng@didiglobal.com
        } \\
}
\maketitle

\begin{abstract}
  This paper explores the potential of curriculum learning in LiDAR-based 3D object detection by proposing a curricular object manipulation (COM) framework.
  The framework embeds the curricular training strategy into both the loss design and the augmentation process.
  For the loss design, we propose the COMLoss to dynamically predict object-level difficulties and emphasize objects of different difficulties based on training stages.
  On top of the widely-used augmentation technique called \textit{GT-Aug} in LiDAR detection tasks, we propose a novel COMAug strategy which first clusters objects in ground-truth database based on well-designed heuristics.
  Group-level difficulties rather than individual ones are then predicted and updated during training for stable results.
  Model performance and generalization capabilities can be improved by sampling and augmenting progressively more difficult objects into the training samples.
 Extensive experiments and ablation studies reveal the superior and generality of the proposed framework.  
 The code is available at \href{CoRP}{https://github.com/ZZY816/COM}.
\end{abstract}
\section{Introduction}\label{sec:intro}

\begin{figure}[t]
  \centering
  \begin{subfigure}{0.235\textwidth}
    \includegraphics[trim={22cm 7cm 19cm 0cm},clip, width=\textwidth]{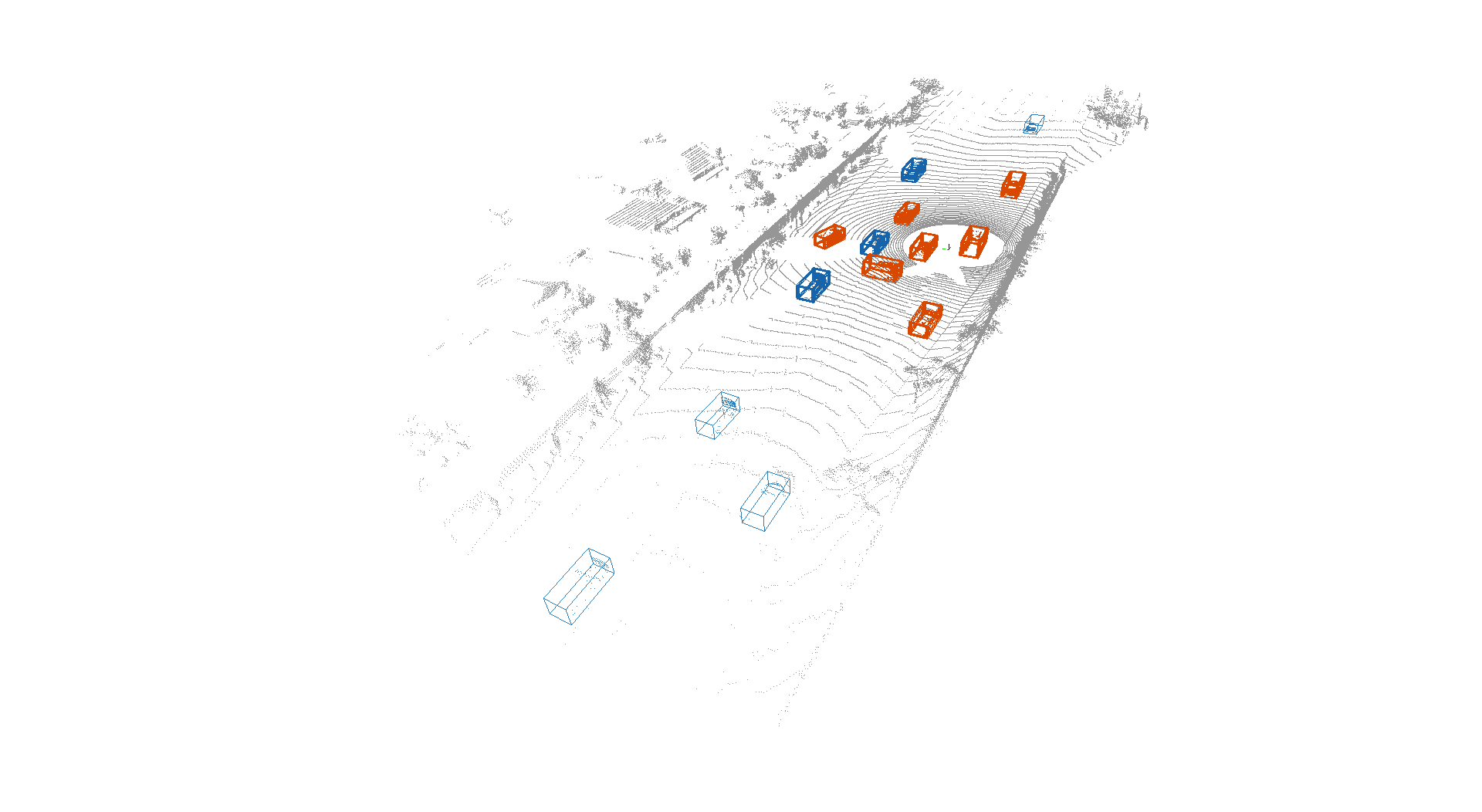}
    \caption{Early stage.}    
      \label{fig:intro1}
    \end{subfigure}
  \begin{subfigure}{0.235\textwidth}
    \includegraphics[trim={21cm 7cm 20cm 0cm},clip, width=\textwidth]{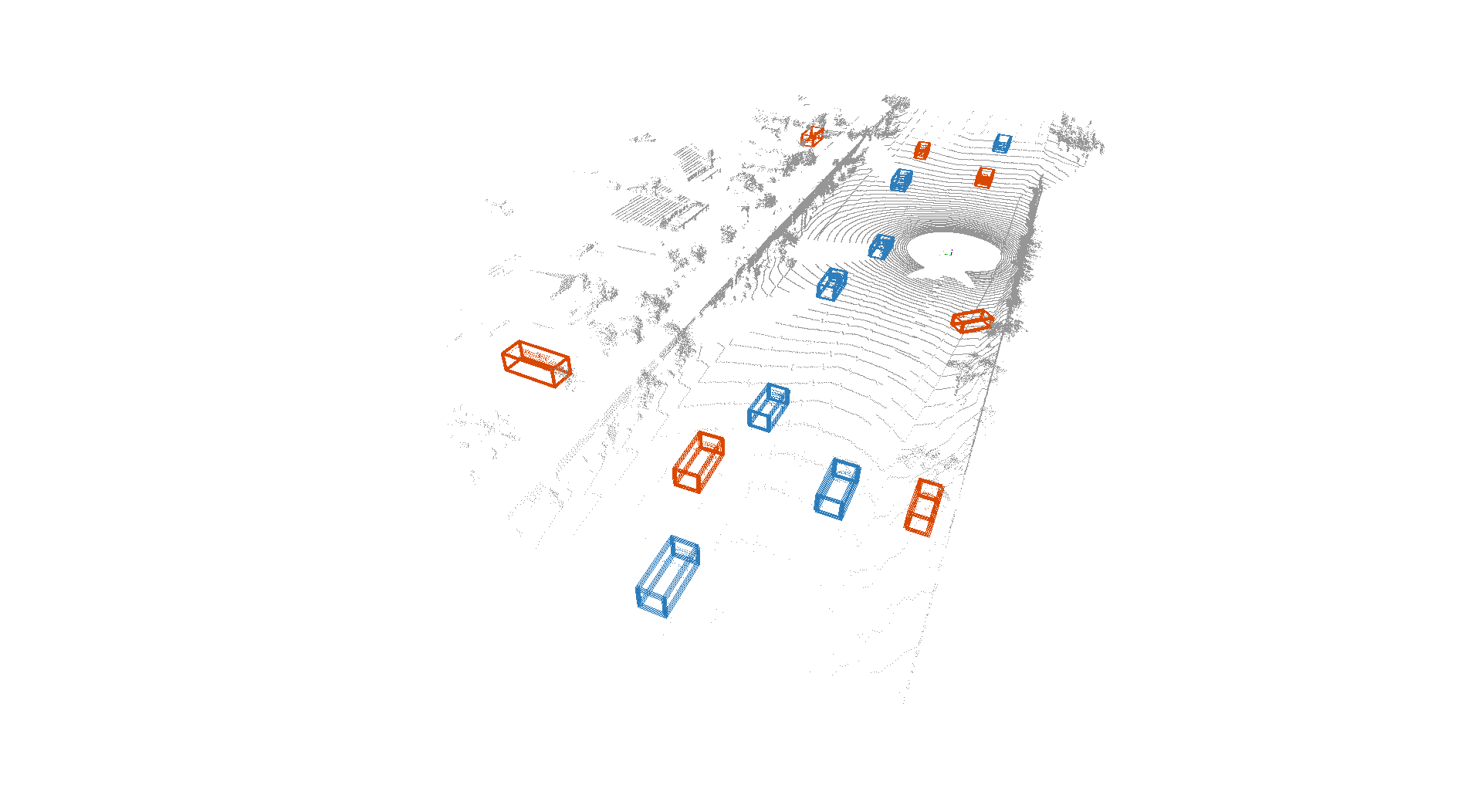}
    \caption{Later stage.}
    \label{fig:intro2}
  \end{subfigure}    
  \caption{
    The proposed Curricular Object Manipulation (COM) works in an easy-to-hard manner.
    In early stages, COMAug constrains the augmented objects (highlighted in red) to be easy ones and COMLoss down-weights losses from difficult objects (marked in boxes with thin lines).
    Objects with varying degrees of difficulty are inserted into the point clouds in later stages.
    On the other hand, hard objects will contribute  more to loss values  as training progresses.
    \textbf{Best viewed in color.}
  } \label{fig:intro}
\end{figure}

LiDAR sensors can provide accurate, high-definition 3D measurements of the surrounding environment. Such 3D information plays a noninterchangeable role in safety-critical applications like 3D object detection in self-driving. 
However, the rich 3D information from LiDAR sensors does not come without problems.
Usually presented in the form of a point cloud, LiDAR data suffers from
\begin{enumerate*}[label=(\roman*)]
\item non-uniformity: the point density decreases monotonically as the laser range increases;
\item orderless: the geometry of a point cloud remains unchanged even if all of its points are randomly shuffled.
\item sparsity: when quantized into voxel grids, a significant portion of the voxels are empty;
\end{enumerate*}

To build a robust and performant LiDAR object detector, different data representations have been explored to alleviate the non-uniformity and orderless challenges.
Feature extraction from the raw orderless point cloud can be made possible by performing radius search or nearest neighbor search in the 3D Euclidean space~\cite{chen2019fast,yang2018ipod,shi2020point,qi2019deep}.
Another popular solution is to quantize the input point cloud into a fixed grid of voxels~\cite{zhou2018voxelnet} or pillars of voxels~\cite{lang2019pointpillars}. At the price of quantization error, later processing can be done efficiently on the regular voxel pillars or grids~\cite{yan2018second}. 

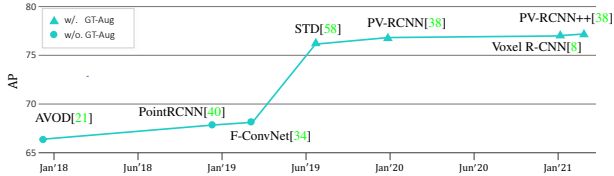
\begin{figure}[t]
\centering
\begin{overpic}[width=0.45\textwidth]{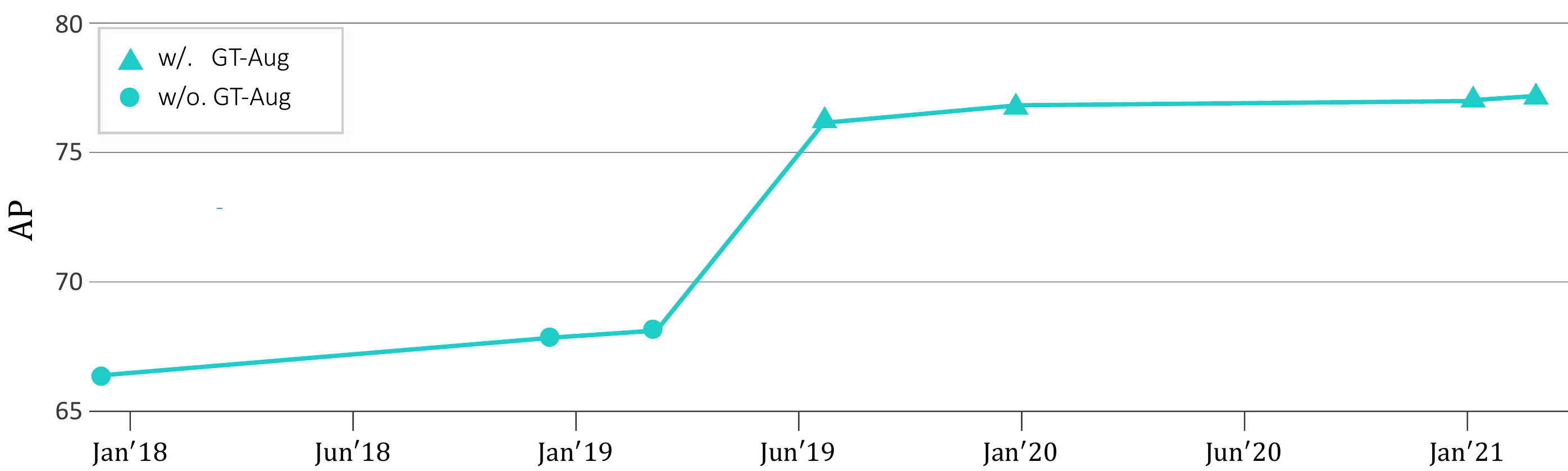}
\put(10,9){\makebox[0pt]{\tiny{AVOD\cite{ku2018joint}}}}
\put(30,10){\makebox[0pt]{\tiny{PointRCNN\cite{shi2019pointrcnn}}}}
\put(45,6){\makebox[0pt]{\tiny{F-ConvNet\cite{qi2018frustum}}}}
\put(53,24){\makebox[0pt]{\tiny{\text{STD}\cite{yang2019std}}}}
\put(68,25){\makebox[0pt]{\tiny{\text{PV-RCNN}\cite{shi2020pv}}}}
\put(90,21){\makebox[0pt]{\tiny{\text{Voxel R-CNN}\cite{deng2021voxel}}}}
\put(95,26){\makebox[0pt]{\tiny{\text{PV-RCNN++}\cite{shi2020pv}}}}
\end{overpic}
    \caption{3D object detection mAP of the car category with hard difficulties on the KITTI dataset from 2018 to 2021.
    It is obvious from the figure that the GT-Aug strategy boosts the KITTI 3D object detection benchmark by a large margin since its inception~\cite{yan2018second}. GT-Aug has since become the de facto augmentation practice in popular open source toolkits~\cite{mmdet3d2020,openpcdet2020}.}
    \label{fig:gtaug}
\end{figure}

But these different data representations do not change the sparsity of the LiDAR point cloud data.
Compared with image object detection tasks, sparse point clouds contain much less input stimuli and positive samples for neural network training, as depicted in Figure~\ref{fig:intro}. Thus, effective data augmentation strategies are critical for faster model convergence and better detection performance~\cite{yan2018second,fang2021lidar,kim2021point,nekrasov2021mix3d,xiao2022polarmix}. Among them, GT-Aug~\cite{yan2018second} (see Figure~\ref{fig:gtaug}) is widely adopted. GT-Aug first aggregates ground truth labels from the training dataset into a database. During training, randomly selected samples from the database are inserted into the point cloud to amplify the supervision signal.

Notice that GT-Aug treats all samples in the database equally, and all epochs of the training process equally. It has brought to our attention that selecting too many hard examples at early stages may overwhelm the training, while selecting too many easy samples at the later stages may slow the model convergence. Similar conclusions were also reached independently in the facial recognition field~\cite{huang2020curricularface}. This important finding raises two questions for the widely used GT-Aug strategy: 
\begin{enumerate*}[label=(\roman*)]
\item at a given training stage, how to select samples that benefit the current stage the most,
\item at different training stages, how to adjust the sampling strategies accordingly.
\end{enumerate*}
However, solving these two questions is not yet enough as the original objects in the training sample can also be ill-suited for current training.
  Therefore, we raise one additional question as
\begin{enumerate*}[label=(\roman*)]
\item[(iii)] how to properly handle augmented objects as well as original objects can contribute to the model performance.
\end{enumerate*}

This work answers the above questions by leveraging curriculum learning.
Curriculum learning draws inspiration from the human cognitive process, which begins with easier concepts and gradually moves on to more complicated ones~\cite{bengio2009curriculum, soviany2022curriculum}.
Enlightened by such easy-to-hard paradigm, we propose a curricular object manipulation (COM) framework for the LiDAR object detection task. Our framework consists of 
\begin{enumerate*}[label=(\roman*)]
\item \textbf{COMLoss} to manipulate the contributions from objects of different difficulties, and
\item \textbf{COMAug} to manipulate the sampling process in GT-Aug.
\end{enumerate*}

In the COM framework, we employ the classification loss as a simple yet effective proxy for object difficulties.
The COMLoss suppresses loss contributions from hard objects in earlier stages and gradually looses the suppression, as depicted in \cref{fig:intro}.
Unfortunately, using classification score as the difficulty proxy 
can cause an inevitable paradox in COMAug.
Specifically, COMAug relies on update-to-date scores of all objects to perform difficulty-adaptive augmentation.
In contrast, all objects should be sampled recently for augmentation to update their scores, which is impossible because of the limited number of augmented objects in each training frame.
We design a clustering based method to address such paradox: objects with similar difficulties are grouped together, and the difficulty estimates are updated for the groups rather than for the individual objects.
During training, hard groups will be sampled with monotonically increasing probabilities as epoch increases, while objects within each group will be sampled uniformly.
In our work, objects are grouped by their geometry attributes, such as distance, dimension, angle, 
and occupancy ratio. 

We demonstrate the efficacy of our proposed method through extensive experiments and ablation studies. In summary, our contributions include:
\begin{itemize}
  \itemsep0em
\item We propose the COM framework which embeds the easy-to-hard training strategy into both loss design and augmentation process in LiDAR-based object detection.
  For the loss design, COMLoss is introduced to dynamically predict object-level difficulties, based on which we emphasize objects to different extents when the training proceeds.
  For the augmentation, a well-designed COMAug first clusters objects in ground-truth database with carefully-picked heuristics.
  During training, COMAug updates group-level difficulties and controls sampling process in augmentation in a curricular manner.
\item To the best of our knowledge, COM is the first to explore the potentials of curriculum learning in conventional LiDAR-based 3D object detection task.
Extensive experiments and ablation studies reveal the superiority and generality of the proposed framework.  
\end{itemize}

\section{Related Works}\label{sec:related_work}

\subsection{LiDAR-based 3D Object Detection}
LiDAR based object detection aims at localizing objects of interest from the input point cloud.
Current works in this area can be roughly classified based on their LiDAR data representation.
Range view based solutions~\cite{fan2021rangedet, bewley2020range, meyer2019lasernet, chai2021point, sun2021rsn} have high computation efficiency due to the compactness of the 2D range view representation, but usually have inferior detection performance caused by the 2D-3D dimensional gaps.
By directly extracting features from raw point clouds, point-based detectors~\cite{chen2019fast, shi2019pointrcnn, li2021lidar, yang2018ipod, zhang2021varifocalnet, qi2018frustum, zhang2022not, yang20203dssd, shi2020point, qi2019deep, shi2020pa2} achieve satisfying performances but commonly suffer from high computational costs incurred by radius search or nearest neighbor queries in the 3D Euclidean space.
In contrast, voxel-based detectors first transform non-uniform point clouds into regular 2D pillars or 3D voxels and employ convolutions for efficient processing in later stages.
Pioneering works, including VoxelNet~\cite{zhou2018voxelnet} and PointPillars~\cite{lang2019pointpillars}, demonstrate great efficiency-utility trade-off, thus attract much attention from the community~\cite{yan2018second,yin2021center,fan2022embracing,duan2019centernet,dosovitskiy2020image}.
Without loss of generality, we primarily focus on voxel-based methods in our experiments.

\subsection{Data Augmentation in Point Clouds}

Constrained by the enormous annotation cost, public LiDAR datasets usually come in with much smaller volume compared with image datasets, \eg, 15K frames in KITTI~\cite{Geiger2012CVPR} compared with 328K images in MS-COCO~\cite{lin2014microsoft}.
Thus, effective data augmentation strategies are critical for the performance and the generalization capabilities of LiDAR object detection models.

In addition to simple geometry deformations, such as \textit{random rotation}, \textit{random flip}, and \textit{translation}, LiDAR tasks usually employ ground-truth augmentation~\cite{yan2018second} to mitigate the sparsity issue in point clouds.
Before training, ground truth objects with their corresponding point clouds are first collected into a database.
During training, additional ground truth objects sampled from the database are concatenated into current training point to supplement the supervision.
This strategy, termed as GT-Aug, has been widely used in current literature~\cite{yan2018second,yang2022graph,yang20203dssd,yin2021center,shi2020pv,shi2021pv} and popular open source toolkits~\cite{mmdet3d2020,openpcdet2020}.
Thus, our experiments are concentrated on the most effective GT-Aug strategy. 

\subsection{Curriculum Learning}
Curriculum learning improves model performance and generalization by progressively drawing harder data samples for training~\cite{bengio2009curriculum, soviany2022curriculum, castells2020superloss, hacohen2019power, ren2018learning, jiang2018mentornet, chang2017active, kong2021adaptive, li2022curriculum}.
In spite of the fact that curriculum learning has demonstrated its effectiveness in certain classification tasks, its application in object detection tasks~\cite{saxena2019data, castells2020superloss}, especially in LiDAR 3D detection~\cite{yang2021st3d}, remains largely unexplored.
In \citet{saxena2019data}, each object is assigned with a difficulty value that is updated by gradient during training.
Superloss~\cite{castells2020superloss} proposes to find close-form solutions on the fly and down-weight contributions of samples with high variance.
In our work, we simply employ the loss value as the indicator of difficulty.
Without loss of generality, the module can be directly replaced by those in works\cite{saxena2019data, castells2020superloss} for better performance.


In addition to training recipe, curriculum learning is also proved to be successful for data augmentation~\cite{lu2022pcc, wei2021few, sawhney2022ciaug,takase2021self}.
Following the same high-level idea of presenting synthetic data with increasing difficulties, we aim at designing a curricular scheme for the GT-Aug in point clouds.

\section{Methodology}\label{sec:method}
\begin{figure*}
  \centering
  \includegraphics[trim={45pt 100pt 90pt 100pt},clip, width=0.95\textwidth]{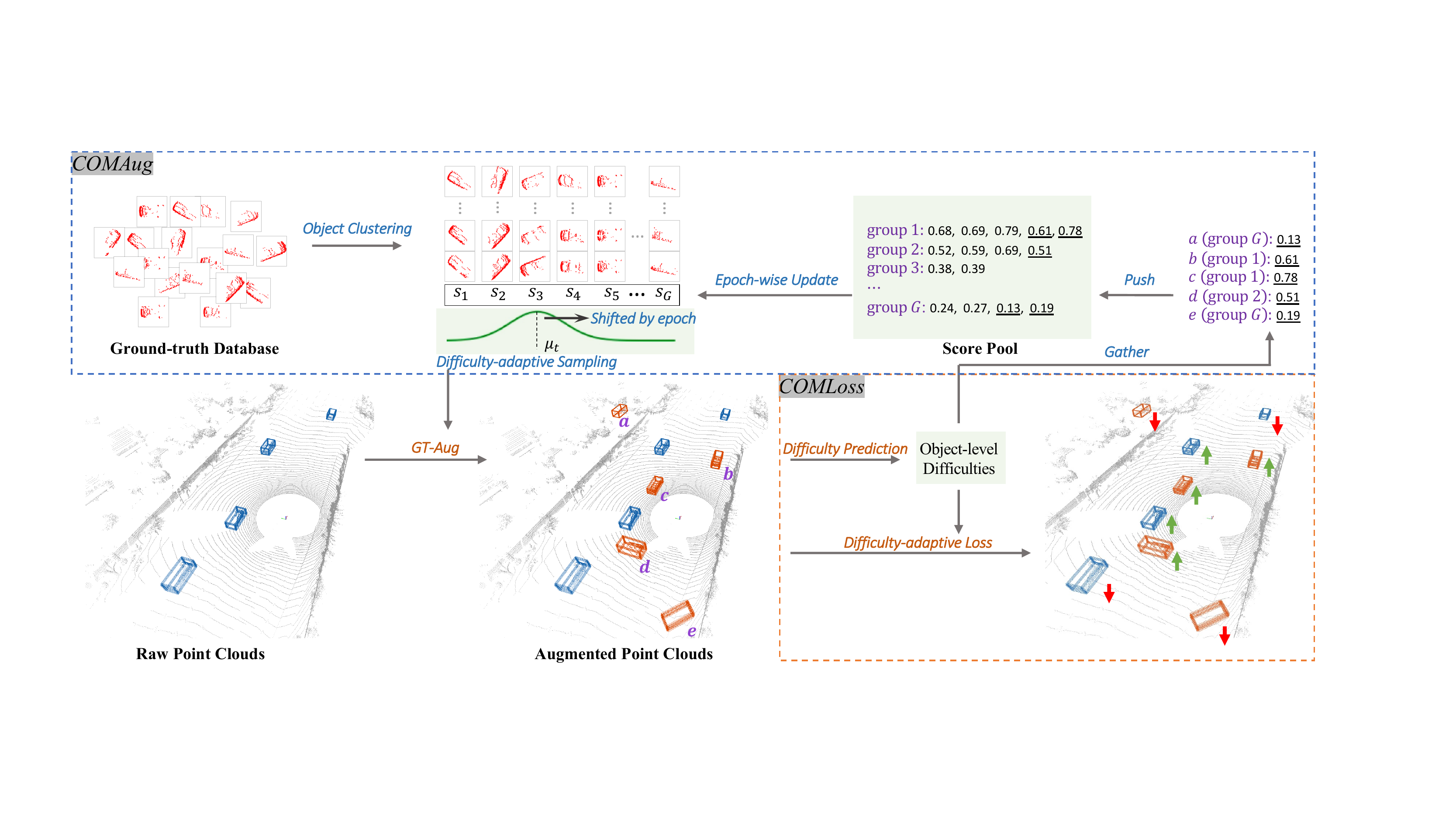}
  \caption{The proposed COM framework is mainly composed of  COMAug and COMLoss.
    (1) COMAug firstly clusters objects in the ground-truth database into $G$ groups by well-designed heuristics.
    At the beginning of each epoch, these groups are assigned with scores $\{s_{g}\}_{1}^{G}$ based on a score pool storing the predicted difficulties of sampled objects in the last epoch.
    According to group scores $\{s_{g}\}_{1}^{G}$, the difficulty-adaptive sampling selects increasingly difficult objects for augmentation as training proceeds, which is realized by a curve with a shifted-by-epoch center $\mu_{t}$ .
    (2) After GT-Aug inserts sampled objects into raw point clouds, COMLoss predicts difficulties for target objects in the augmented point clouds.
    Then object losses are re-weighted (marked by up/down arrows) with respect to object difficulties and training stages.    
  }
  \label{fig:method}
\end{figure*}

In this section, we present details of the curriculum object manipulation (COM) framework, as illustrated in \cref{fig:method}. Basic concepts and notations in LiDAR-based object detection are first reviewed in \cref{subsec:method1}. COMLoss, which manipulates the losses from objects of different difficulties, is described
in \cref{subsec:methodloss}. COMAug, which manipulates the object sampling process in GT-Aug, is described in \cref{subsec:methodaug}.


\subsection{Notations and Definitions}\label{subsec:method1}
The GT-Aug strategy firstly collects a database $\mathset{B} = \{(x_i, y_i)\}_{1}^{n}$ of ground-truth objects. 
For each ground-truth object, 3D coordinates and additional features (such as intensity, elongation, \etc) for 3D points belonging to the object are collected as $x_i$, while human annotations (such as bounding box center position, dimension, yaw angle, category, \etc) are collected as $y_i$.
At any given training sample, assuming there are $P$ annotated ground-truth objects in the current LiDAR point cloud, the GT-Aug strategy works by complementing the target objects up to a predefined threshold $\Gamma$ by sampling $Q = \min(0, \Gamma - P)$ objects from the pre-collected database $\mathset{B}$.

Given the $P$ original objects and the $Q$ augmented objects, the optimization objective for LiDAR based 3D object detectors can be summarized into the general form below:
\begin{equation}  
  \label{eq:general_loss}
    \mathcal{L}  = \frac{1}{N} \left( \mathcal{L}_{n} + \sum_{p=1}^{P}(\mathcal{L}_{c}^{p} + \mathcal{L}_{r}^{p}) +  \sum_{q=1}^{Q}(\mathcal{L}_{c}^{q} + \mathcal{L}_{r}^{q}) \right)      
\end{equation}
For original objects $p$ and augmented objects $q$, their classification loss $\mathcal{L}_{c}$ and regression loss $\mathcal{L}_{r}$ are both taken into consideration during training. $\mathcal{L}_{n}$ is introduced to handle the point cloud background. The normalization factor $N$, which depends on the number of positive labels, is usually customized by different detection models.

\subsection{COMLoss}\label{subsec:methodloss}
The first component of the COM framework is the COMLoss aiming at improving the model performance and generalization capabilities by the curriculum learning scheme.
Despite the demonstrated success of curriculum learning in classification tasks, applying the curriculum learning paradigm in the object detection context has rarely been researched~\cite{saxena2019data,castells2020superloss}.
There are two potential causes of the situation.
Firstly, the curriculum learning framework treats one image sample indivisibly and performs sample-wise manipulation for classification tasks.
In contrast, one sample (image or point cloud) in the object detection context typically contains objects of varying difficulties, necessitating object-level difficulty predictions and pacing patterns.
Secondly, modern object detectors usually have some hard-mining strategies built-in, for example, through
OHEM~\cite{shrivastava2016training}, or focal loss~\cite{lin2017focal} and its variants~\cite{zhang2021varifocalnet,li2020generalized}.
Applying curriculum learning into object detection tasks needs
careful balance between the easy-to-hard curriculum learning paradigm, and the built-in hard-mining scheme.

\subsubsection{Difficulty Criterion}
LiDAR-based object detectors usually assign a categorical likelihood for each predicted object of interest.
Following the idea that loss values can be used as alternative indicators~\cite{kong2021adaptive,huang2020curricularface,meng2021magface,ren2018learning}, we collect the predicted classification scores to gauge the object-level difficulties.
Alternative solutions of object-level difficulties estimation include using handcrafted heuristics~\cite{soviany2021curriculum} and pre-trained models~\cite{ren2018learning,lu2022pcc}. Compared with these, our choice of using the classification scores as the difficulty criterion has clear advantages:
\begin{enumerate*}[label=(\roman*)]
\item classification scores are evaluated on-the-fly as the model being trained, thus these scores align much better with the model optimization status quo, compared with pre-trained models or heuristics;
\item classification scores are evaluated as part of the forward loss computation, thus no additional computation costs are incurred.
\end{enumerate*}

However, there still exists statistical divergence between the classification score distribution and the object difficulty distribution. Such divergence also changes as training progresses. 
For example, scores of easy objects tend to be small at early training stages, while hard object scores can be relatively large in later stages.
Using $\{s_{p}\}_{1}^{P}, \{s_{q}\}_{1}^{Q}$ to represent the collected classification scores for current original and augmented objects, we mitigate the discrepancies by introducing an adaptive threshold $\tau$ which is essentially the running average of the original object scores:
\begin{equation}
  \label{eq:comloss_threshold}
  \tau \leftarrow (1-\alpha)\cdot \tau + \alpha \cdot (\sum_{p}s_{p})/P, 
\end{equation}
where $\tau$ is initialized as 0 and $\alpha$ is the momentum coefficient. We do not include scores of the augmented objects $q_i$ when estimating $\tau$ as augmented objects can have fluctuating scores due to the sampling process randomness.
Given adaptive threshold $\tau$, object difficulties are represented by:
\begin{equation}
  \label{eq:comloss_difficulties}
  \begin{split}
  \tilde{s}_{p} = s_{p} - \tau,\quad \forall p \in \{1, 2, \cdots, P\},\\
  \tilde{s}_{q} = s_{q} - \tau, \quad \forall q \in \{1, 2, \cdots, Q\}.
  \end{split}
\end{equation}

It's noteworthy to mention that a small $\tilde{s}_p$ corresponds to hard samples.
Although this is slightly counterintuitive, we maintain it for the sake of clarity in the follow-up sections.


\subsubsection{Difficulty-adaptive Loss}
The goal of COMLoss is to formulate a mechanism to divert the training focus to different objects at different stages of model training. In particular, in earlier stages easier objects should be emphasized while in later stages harder objects should attract more attention.
To reach this goal, we design an adaptive weighting function $w$ to dynamically adjust the focus of the optimization objective as training proceeds:
\begin{equation}
  \label{eq:comloss_height}
  \begin{split}
    w &= 1 + h_{t}\cdot (1 - e^{\beta\cdot\tilde{s}})/(1+e^{\beta\cdot\tilde{s}}), \\
    & \text{   where   } h_{t} = H\cdot (t_{r}-t)/T.
  \end{split}
\end{equation}
Here $\tilde{s}$ is the object difficulty and $\beta$ controls the curve shape.
$h_{t}$ is the height of the Sigmoid curve at epoch $t$ and $T$ is the number of total epochs.
$H$ is a parameter to control re-weighting degree.
Before epoch $t_{r}$, the $h_{t}$ is greater than 0 and the easy/hard samples are emphasized/suppressed respectively.
For epochs $t>t_{r}$, hard samples are emphasized while easy samples become down-weighted.
We call the epoch $t_{r}$ as the tipping point which is an important parameter for COMLoss.
We found through experiments that the easy-to-hard strategy can still bring performance gains even after setting $t_r$ very close to $T$.
When $t$ approaches $t_r \approx T$, the weighting function $w$ becomes almost flat, thus treating all objects equally. We believe that the built-in hard example mining mechanisms (such as focal loss~\cite{lin2017focal,zhang2021varifocalnet,li2020generalized}) lead to the performance gains in such scenarios.

Given the weighting function $w$, the proposed COMLoss formulation becomes:

\begin{equation}
  \small
    \mathcal{L}  = \frac{1}{N}\left(\mathcal{L}_{n} + \sum_{p=1}^{P}w_{p}\cdot(\mathcal{L}_{c}^{p} + \mathcal{L}_{r}^{p}) +  \sum_{q=1}^{Q}w_{q}\cdot(\mathcal{L}_{c}^{q} + \mathcal{L}_{r}^{q})\right).
\end{equation}

\subsection{COMAug}\label{subsec:methodaug}
During the training of LiDAR-based 3D object detectors, GT-Aug randomly samples objects from the ground-truth database $\mathcal{B}$ and inserts them into current training point.
Nevertheless, GT-Aug cannot foretell whether the sampled objects would benefit the training more than the left out ones.
Intuitively, sampling harder samples in early stages can increase learning difficulty and consequently lead to unstable training.
On the other hand, easier samples will not bring in much useful knowledge in the later stages.

One solution to this problem is the curricular scheme which samples increasingly difficult objects for augmentation.
But known and up-to-date object difficulty scores are a prerequisite for such curricular scheme.
To fulfill such requirement, a naive way is to initialize object difficulties with the same score value and update an object's difficulty if only sampled for augmentation.
This solution however has two significant flaws:
(1) The probability of one object to be sampled is small because of the enormous volume of the ground-truth database. Thus, object difficulties are mostly not updated throughout training, or have predictions that are severely outdated.
For example, the Waymo Open Dataset~\cite{sun2020scalability} contains 12.6 million 3D bounding box annotations in LiDAR data. The probability of one vehicle being sampled for augmentation during one epoch of training is less than 1\%.
(2) The sampling process can be easily trapped in a downward spiral: objects have limited chances of being selected for augmentation, thus their difficulty estimations can be inaccurate. 
Improperly sampled augmentation objects may introduce perturbations in model training, which further hinders the reliability of the updated difficulty estimations.
These issues motivate the COMAug to cluster objects in the database (\cref{sec:comaug_oc}) and update group-level scores rather than individual ones (\cref{sec:comaug_score}).
By design, COMAug is able to perform difficulty-adaptive sampling (\cref{sec:comaug_sampling}), achieving the curricular augmentation along the training process.

\subsubsection{Object Clustering} \label{sec:comaug_oc}

The core of COMAug is to assign objects with similar properties to the same cluster.
Although point cloud provides rich geometry information for objects of interest, how to select effective grouping criterion is critical for COMAug.
For example, extensive works have demonstrated that object distances~\cite{shi2020pv, fan2022embracing, yang2022graph, deng2021voxel, mao2021voxel}, occlusion~\cite{xu2021behind}, object sizes and angles~\cite{yin2021center} are all influential factors for detection quality.

\begin{figure}[htb!]
  \centering
    \includegraphics[trim={22cm 6cm 20cm 6cm},clip, width=0.4\textwidth]{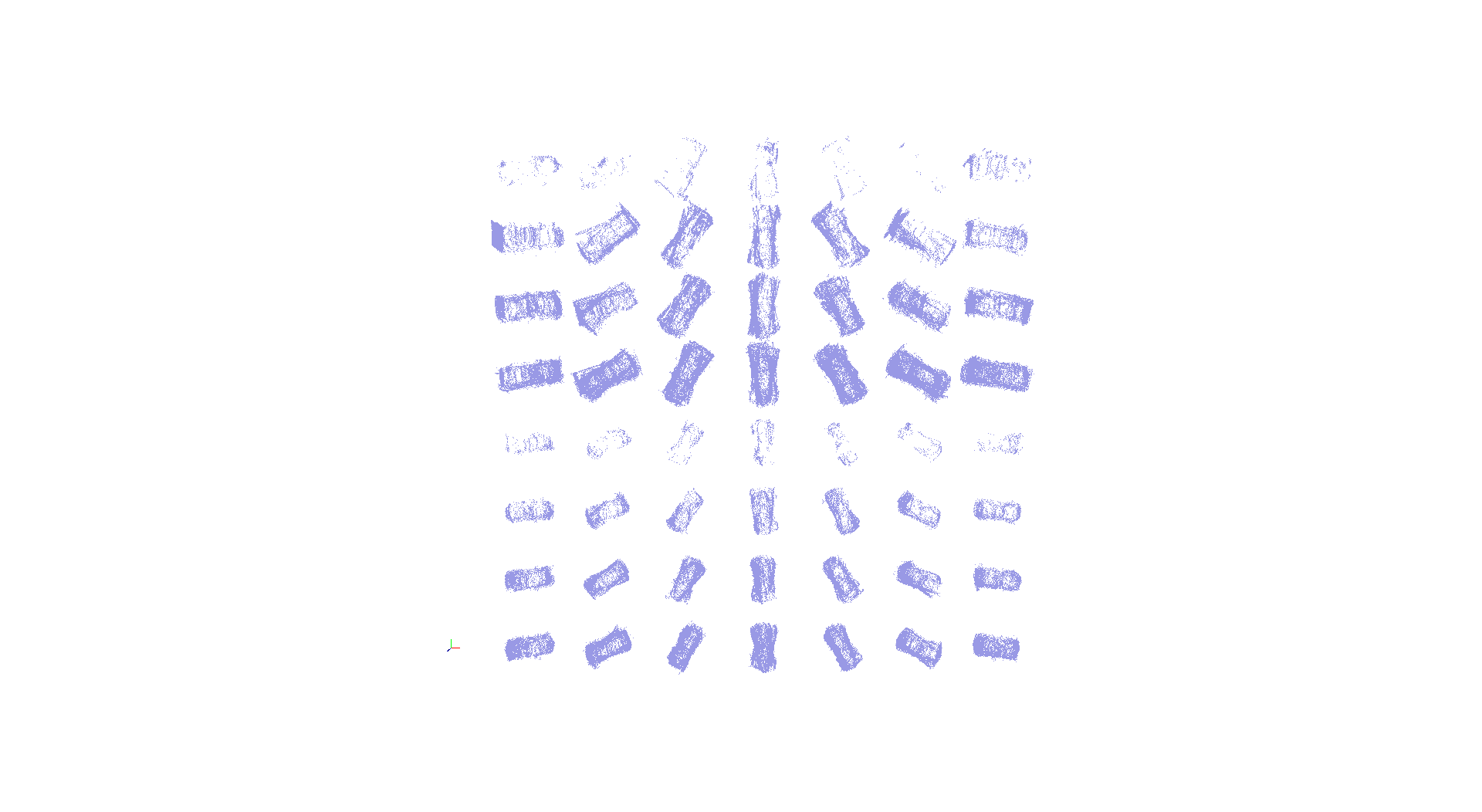}  
  \caption{
    Visualization of several groups of the vehicle class.
    Each sub-graph shows the aggregation of point clouds from 20 random objects in one group.
    Top four rows show vehicles with sizes $f_s \in [5, 8]$ meters, while bottom four rows shows vehicles with sizes no larger than 5 meters.
    In each four rows, the occupancy ratio $f_{o}$ increases from top to bottom.
    Changes of angle $f_{a}$ can be observed in the left-to-right direction.
  }
  \label{fig:cluster}
\end{figure}

Assuming the bounding box for one object is $[x, y, z, l, w, h, \alpha]$, where $x, y, z$ are the coordinates of the box center.
$l, w, h, \alpha$ are the length, width, height and heading of the box.
Drawing inspirations from the literature and our empirical studies, the following four factors are used to perform object clustering in COMAug:
(1) the distance $f_{d}$ to the LiDAR sensor calculated by $f_{d} = \sqrt{x^{2}+y^{2}+z^{2}}$;
(2) the size $f_{s}$ of the bounding box represented by $f_{s}=\max(l, w, h)$; 
(3) the relative angle $f_{a}$ between the box heading and the azimuth of the box center, \ie, $f_{a} = \alpha - \arctan(y/x)$; 
and (4) the occupancy ratio $f_{o}$ of observed area over the entire bounding box.
In our implementation, we divide the object into a particular number of equally spaced 3D voxels and assign the ratio of non-empty ones to $f_o$.
\cref{fig:cluster} shows a visualization of clustering results from our experiments.

\vspace{-15pt}

\subsubsection{Group-level Score Update}\label{sec:comaug_score}
We group the aggregated ground truth objects into $g$ groups and each group has a difficulty score $\{s_{g}\}_{1}^{G}$. At any given moment during training, difficulty scores of the augmented objects (\ie, $\{\tilde{s}_{q}\}_{1}^{Q}$), as predicted from the COMLoss, are also gathered.
For each group, we design a score pool $\mathcal{P}_g$ which stores all the scores gathered during one epoch of training (see \cref{fig:method}).
At the end of each epoch, the score of each group $s_g$ is updated to the average of the score pool $\mathcal{P}_g$.
Compared with the aforementioned individual-level update paradigm, the group-level score update can greasy lessen the issues of outdated and unstable score predictions.

Also, we found that for group score $s_g$, epoch-wise update works better than momentum update in COMLoss.
It has brought to our attention that the distribution of group score pool sizes ${|\mathcal{P}_g|}$ can be highly skewed during training. For example, easy groups can have a much larger score pool, especially in early stages. Momentum update thus will make groups update in different paces, hindering the effectiveness of carefully designed curriculum learning paradigm. Epoch-wise update, on the other hand, can circumvent the issue.

\subsubsection{Difficulty-adaptive Sampling}\label{sec:comaug_sampling}
The last component of COMAug is a difficulty-adaptive sampler.
Given the sampler, for each query, we first select one group, and then randomly sample one object from the chosen group.
The sampler should further fulfill these requirements: (1) adaptive to difficulty scores; (2) adaptive to training stages; and (3) robust to the clustering results.

We introduce a Gaussian sampling curve centered at $\mu_{t}$ with hyperparameter $\sigma$, where $\mu_{t}$ shifts with epoch number $t$.
With this curve, each group $g$ is assigned with a probability $p_{g} = \exp\{-(\tilde{s}_{g}-\mu_t)^2/(2\sigma^2)\}$.
By this means, the sampling is adaptive to difficulties as groups with scores close to $\mu_{t}$ are of high chances to be selected.
We consequently make $\mu_{t}$ to change by epoch $t$ to meet the second requirement.
Without loss of generality, we assume the group scores $\{s_{i}\}_{1}^{G}$ are already sorted in descending order.
Then we assign $\mu_{t}$ with the score from harder groups as epoch $t$ increases.
Specifically, we let
\begin{equation}
  \label{eq:comaug_mu}
  \mu_{t} = s_{g}, \text{ where } g = \min (\lfloor(\lambda\cdot t/T)\cdot G\rfloor, G),
\end{equation}
where $T$ is the number of total epochs and $\lambda$ is a hyperparameter controlling the pacing speed.

Directly using the probability $p_{g}$ to sample the groups can be problematic when the clustered groups have highly imbalanced numbers of elements.
For example, with the same group difficulties, objects in tiny groups can be repeatedly sampled 
while those in the large group will be under-sampled.
Therefore, we propose to normalize the $p_{g}$ by making summations of all probabilities to be 1 as well as considering the group size $n_{g}$.
In the end, the probability of selecting group $g$ is computed as:
\begin{equation}
  \label{eq:comaug_samplingprob}
  p^{g} = (p_{g}\cdot n_{g})/(\sum_{i=1}^{G}p_{i}\cdot n_{i}).
\end{equation}



\section{Experiments}\label{sec:exp}

In this section, we detail the experimental setup in \cref{sec:exp_setup} and present the main results in \cref{sec:exp_main}.
Comprehensive ablation studies on the proposed COMLoss and COMAug are shown in \cref{sec:ablation_loss} and \cref{sec:ablation_aug}, respectively. 

\subsection{Experimental Setup}\label{sec:exp_setup}
\noindent\textbf{Implementation Details.}
We implement our method on top of OpenPCDet~\cite{openpcdet2020} and use the default configurations.
Specifically, we train all detectors for 30 epochs by an Adam one cycle optimizer.
The learning rate is warmed up from 0.003 to 0.03 within the first 12 epochs before dropping.
Experiments are conducted on 8 A100 GPUs and batch size in each GPU is 2 for CenterPoint~\cite{yin2021center}, PointPillars~\cite{lang2019pointpillars} and SECOND~\cite{yan2018second}.
Note here we use the one-stage and pillar-based version of CenterPoint.
We fix random seeds for fair comparisons.
For GT-Aug, the thresholds $\Gamma$ for class vehicle, pedestrian and cyclist are 15, 10, 10 respectively.

\noindent\textbf{Dataset and Metrics.}
Our experiments are conducted on the large-scale Waymo Open Dataset~\cite{sun2020scalability} (WOD) which contains 798 sequences for training, 202 for validation, and 150 for testing respectively.
20\% training data (32k frames) are uniformly sampled for training.
For evaluation, we report the 3D LEVEL\_1 average precision (AP$_{L1}$) and LEVEL\_2 average precision (AP$_{L2}$) on the WOD validation set, where LEVEL\_1 and LEVEL\_2 represent the level of difficulty.

\noindent\textbf{COM Configurations.}
Unless stated otherwise, the parameters of COMLoss are default to $\alpha=0.001$ for \cref{eq:comloss_threshold} and $t_{r}=30, \beta=-5$ for \cref{eq:comloss_height}.
$H$ in \cref{eq:comloss_height} is the hyperparameter in experiments.
We let $\lambda=0.5$ and $\sigma=0.2$ in the sampling curve in COMAug.
Details of object clustering are placed in the supplementary due to the limited space.

\begin{table}[t!]
  \centering
\renewcommand{\arraystretch}{1.05}
\renewcommand{\tabcolsep}{2pt}
  \small
\begin{tabular}{llll}
  \hline\toprule
  \multirow{2}{*}{Detector} & \multirow{2}{*}{Method} & \multicolumn{2}{c}{Pedestrian}  \\
        \cline{3-4}  
    &  &  \multicolumn{1}{c}{3D AP$_{L1}$ (\%)} & \multicolumn{1}{c}{3D AP$_{L2}$ (\%)}\\ 
  \hline
  \multirow{4}{*}{SECOND~\cite{yan2018second}}
                                  & + \textitf{None}	 & 62.37 & 54.05  \\
   \cline{2-4}  
                                  & + \textitf{GT-Aug}	 & 65.40 \improveb{3.03} & 57.07   \improveb{3.02}  \\
                                  & + \textitf{COMLoss}	 & 66.23 \improveb{3.86} & 57.77 \improveb{3.72} \\
                                  & + \textitf{COMAug} 	 & 66.46 \improveb{4.09} & 57.97 \improveb{3.92} \\
                                  & + \textitf{COM} 	 & 66.66 \improveb{4.29} & 58.14 \improveb{4.09} \\
  \hline
  \multirow{4}{*}{PointPillars~\cite{lang2019pointpillars}}
                                  & + \textitf{None}	 & 64.36 & 55.98  \\
   \cline{2-4}      
                                  & + \textitf{GT-Aug}	 & 65.54 \improvea{1.18} & 57.20 \improvea{1.22}    \\
                                  & + \textitf{COMLoss}	 & 66.19 \improvea{1.83} & 57.80 \improvea{1.82} \\
                                  & + \textitf{COMAug} 	 & 66.75 \improveb{2.39} & 58.40 \improveb{2.42} \\
                                  & + \textitf{COM}       & 66.81 \improveb{2.45} & 58.52 \improveb{2.54} \\
  \hline
  \multirow{4}{*}{CenterPoint~\cite{yin2021center}}
                                  & + \textitf{None}	 & 72.79 & 65.02 \\
  \cline{2-4}
                                  & + \textitf{GT-Aug}	 & 73.62 \improvea{0.83} & 65.84 \improvea{0.82} \\  
                                  & + \textitf{COMLoss}	 & 74.16 \improvea{1.37} & 66.36 \improvea{1.34} \\
                                  & + \textitf{COMAug} 	 & 75.34 \improveb{2.55} & 67.66 \improveb{2.64} \\
                                  & + \textitf{COM}       & 75.43 \improveb{2.64} & 67.76 \improveb{2.74} \\
  \bottomrule
\end{tabular}

\caption{Performances on class pedestrian.
  In green are the gaps of at least +2.0 point.
}\label{tab:main_ped}
\end{table}

\begin{table}[t!]
  \centering
\renewcommand{\arraystretch}{1.05}
\renewcommand{\tabcolsep}{2pt}
  \small
\begin{tabular}{llll}
  \hline\toprule
  \multirow{2}{*}{Detector} & \multirow{2}{*}{Method} & \multicolumn{2}{c}{Vehicle}  \\
        \cline{3-4}  
    &  & \multicolumn{1}{c}{3D AP$_{L1}$ (\%)} & \multicolumn{1}{c}{3D AP$_{L2}$ (\%)}\\ 
  \hline
  \multirow{4}{*}{SECOND~\cite{yan2018second}}
                                  & + \textitf{None}	 & 71.72 & 63.26  \\
   \cline{2-4}  
                                  & + \textitf{GT-Aug}	 & 72.40 \improvea{0.68} & 63.95 \improvea{0.69}    \\
                                  & + \textitf{COMLoss}	 & 72.82 \improveb{1.10} & 64.35 \improveb{1.09} \\
                                  & + \textitf{COMAug}  & 73.07 \improveb{1.35} & 64.57 \improveb{1.31} \\
                                  & + \textitf{COM} 	 & 73.32 \improveb{1.60} & 65.34 \improveb{2.08} \\
                                  
  \hline
  \multirow{4}{*}{PointPillars~\cite{lang2019pointpillars}}
                                  & + \textitf{None}	 & 69.28 & 61.04  \\
   \cline{2-4}      
                                  & + \textitf{GT-Aug}	 & 70.00 \improvea{0.72} & 61.70 \improvea{0.66}     \\
                                  & + \textitf{COMLoss}	 & 70.78 \improveb{1.50} & 62.43 \improveb{1.39}  \\
                                  & + \textitf{COMAug} 	 & 71.01 \improveb{1.73} & 62.88 \improveb{1.84} \\
                                  & + \textitf{COM}       & 71.38 \improveb{2.10} & 62.96 \improveb{1.92} \\
  \hline
  \multirow{4}{*}{CenterPoint~\cite{yin2021center}}
                                  & + \textitf{None}	 & 70.97 & 62.54  \\
   \cline{2-4}      
                                  & + \textitf{GT-Aug}	 & 71.05 \improvea{0.08} & 62.66 \improvea{0.12}    \\
                                  & + \textitf{COMLoss}	 & 71.57 \improvea{0.60} & 63.13 \improvea{0.59} \\
                                  & + \textitf{COMAug} 	 & 72.03 \improveb{1.06} & 63.61 \improveb{1.07}\\
                                  & + \textitf{COM}       & 72.15 \improveb{1.18} & 64.29 \improveb{1.75} \\
  \bottomrule
\end{tabular}
\caption{Performances on class vehicle.
    In green are the gaps of at least +1.0 point.
}\label{tab:main_veh}
\end{table}

\subsection{Main Results}\label{sec:exp_main}

In this part, we show the efficacy and generality of the proposed COM framework on three popular LiDAR-based 3D object detectors: SECOND~\cite{yan2018second}, PointPillars\cite{lang2019pointpillars} and 
CenterPoint~\cite{yin2021center}.
We compare the performances of these detectors when using GT-Aug (denoted as \textit{+GT-Aug}), using COMLoss and GT-Aug (denoted as \textit{+COMLoss}), using COMAug (denoted as \textit{+COMAug}), using both COMLoss and COMAug (denoted as \textit{+COM}), and using vanilla configurations (denoted as \textit{+None}).
\cref{tab:main_ped} and \cref{tab:main_veh} present the results on the pedestrian and vehicle categories.

In \cref{tab:main_ped}, the improvements brought by GT-Aug on AP$_{L1}$ are 3.03\%, 1.18\% and 0.83\% for the three detectors.
COMLoss further increases these values into 3.86\%, 1.83\% and 1.37\%.
Finally, our COM improves the baseline detectors by 4.09\%, 2.54\%, 2.74\% on AP$_{L2}$, respectively.
In \cref{tab:main_veh},
on strongest CenterPoint detector COMLoss and COMAug lead to the improvements of 0.60\% and 1.06\%, respectively.
The COM framework promotes the AP$_{L1}$ to 72.15, which beats the baseline value of 70.97\% by 1.18\%.



\subsection{Ablation Studies on COMLoss} \label{sec:ablation_loss}

This part examines three key parameters for COMLoss in \cref{eq:comloss_height}: tipping point $t_{r}$, parameters $H$, and $\beta$.
Using the CenterPoint~\cite{yin2021center} with GT-Aug as the baseline, we report the improvements on 3D AP$_{L1}$ brought by our COMLoss.

\begin{figure}[htb!]
  \centering
  {\includegraphics[width=1\linewidth]{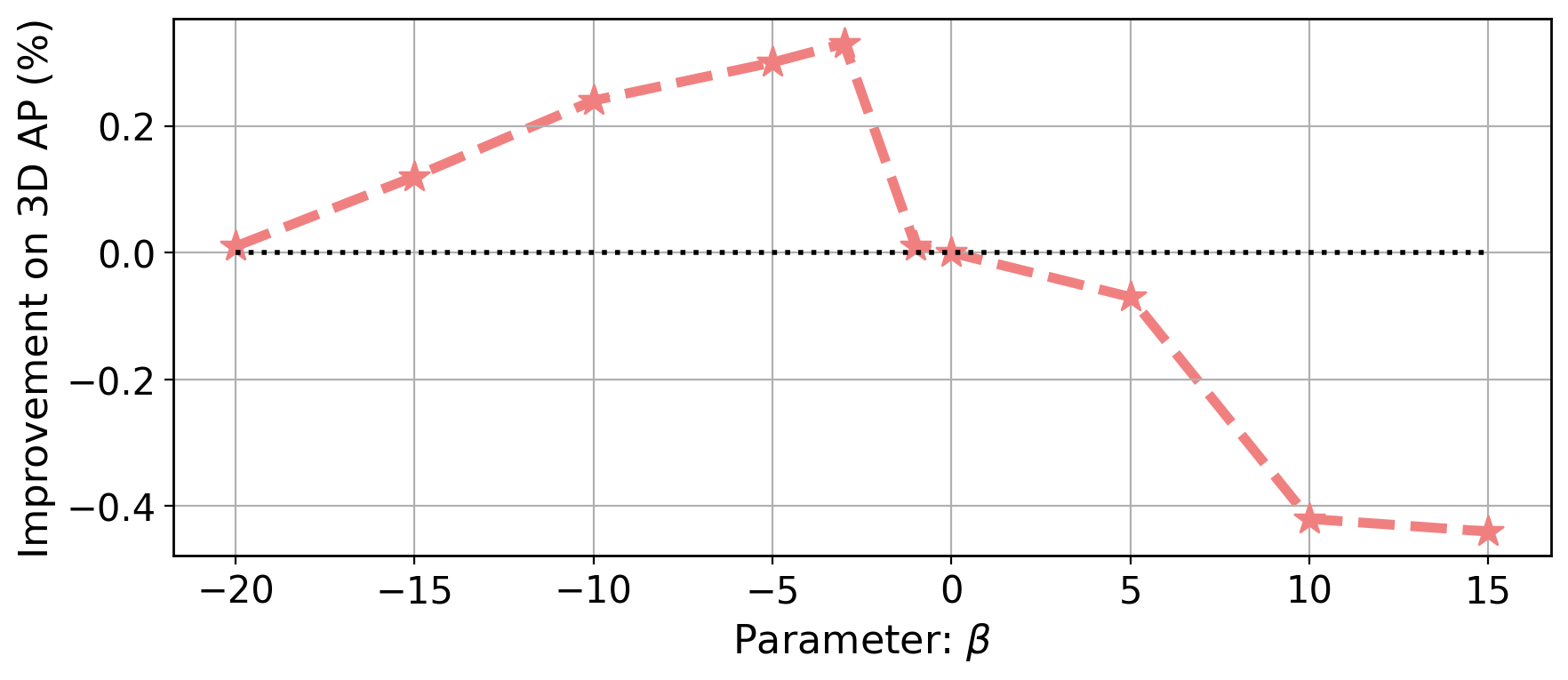}}
  \caption{Effects of parameter $\beta$ on vehicle ($H=1$).} \label{fig:comloss2}
\end{figure}

\noindent
\textbf{Effects of parameter $\boldsymbol \beta$.}
$\beta=0$ degrades model to the baseline and $\beta < 0$ leads to the traditional easy-to-hard strategy.
The reverse-curricular scheme is achieved using $\beta > 0$, where the model emphasizes hard samples at early stages and easy samples at later stages.
On the other hand, decreasing the absolute value $|\beta|$ flattens the curve.
We conduct experiments with $\beta$ ranging from -20 to 15, and present our results in  \cref{fig:comloss2}. 
We find that the reverse-curricular scheme causes performance degradation, while steeper curve worsens the situation.
In contrast, the easy-to-hard scheme can benefit the detector and $\beta=-3$ achieves the best result.

\begin{figure}[htb!]
  \centering
  {\includegraphics[width=1\linewidth]{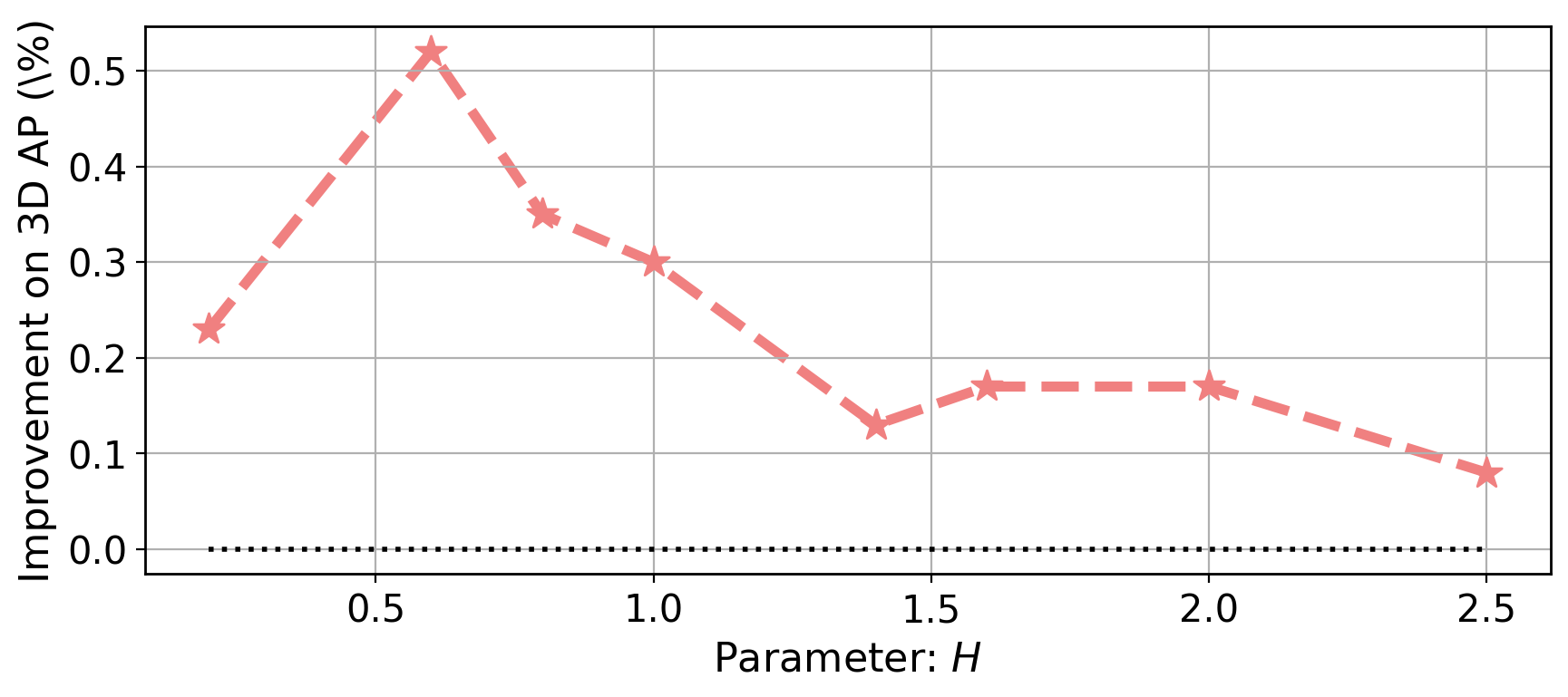}}
  \caption{Effects of parameter $H$ on vehicle.} \label{fig:comloss3}
\end{figure}

\noindent
\textbf{Effects of parameter $\boldsymbol H$.}
\cref{fig:comloss3} shows the result of the ablation study on the parameter $H$, which essentially reflects the extent of re-weighting objects.
We observe that the improvements brought by COMLoss increase first and then decrease as $H$ grows larger, and the maximum performance gain ($>0.5\%$) is obtained at $H=0.6$.


\begin{figure}[htb!]
  \centering
  {\includegraphics[width=1\linewidth]{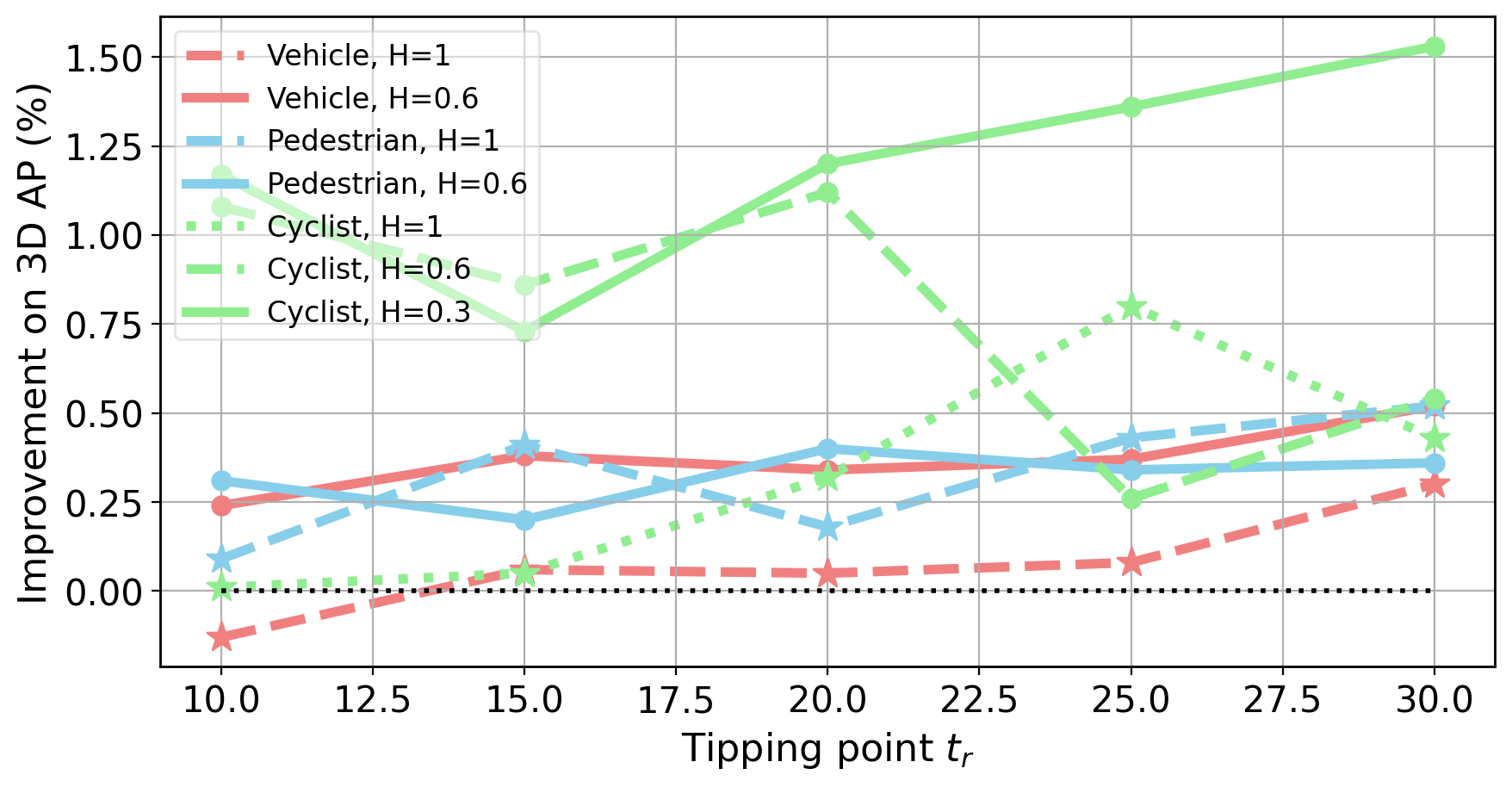}}
  \caption{Effects of tipping point $t_{r}$.
    \textbf{Best viewed in color.}
  } \label{fig:comloss1}
\end{figure}

\noindent
\textbf{Effects of tipping point $\boldsymbol t_r$.}
COMLoss emphasizes easy and hard objects correspondingly before and after the epoch $t_r$.
This is the most crucial parameter in COMLoss as it strikes a balance between the introduced easy-to-hard scheme and the built-in hard-mining strategy in detector.
We conduct extensive experiments on $t_{r}$ with various height $H$ across multiple object categories, and present the results in \cref{fig:comloss1}.
For the class of vehicle, the performances increase with $t_{r}$ for both $H=1$ and $H=0.6$.
The classes of pedestrian and cyclist observe different optimal $t_{r}$ in different cases.
For example, the best $t_{r}$ on pedestrian is $20$ for $H=0.6$ and $30$ for $H=1$.
On cyclist, the best $t_{r}$ is $30, 20, 25$ when $H$ is 0.3, 0.6 and 1, respectively.
In each situation, the optimal $t_{r}$ is near the end of the training.
Additionally, the largest improvements are achieved with different $H$s (\ie, $H=0.6, 1, 0.3$ for vehicle, pedestrian and cyclist) but with the same $t_{r}$ of 30.
This situation may be caused by the built-in hard-mining scheme in detectors.
Setting $t_r$ as 30 means that the re-weighting curve treats all objects equally in the end.
However, hard objects are still been emphasized because of the built-in mechanism.
The whole training therefore still follows the easy-to-hard principle.



\vspace{-10pt}

\subsection{Analysis on COMAug} \label{sec:ablation_aug}
This part presents our analysis on COMAug including the ablation study on the object clustering as well as several visualizations.
All experiments in this part are conducted with CenterPoint~\cite{yin2021center} on the vehicle category from the Waymo Open Dataset.

\noindent
\textbf{Effects of clustering factors.}
COMAug clusters objects by four factors including distance $f_{d}$, size $f_{s}$, angle $f_{a}$ and occupancy ratio $f_{o}$.
We examine the performances of COMAug with various combinations of these factors, and report the results in \cref{tab:clusterfactor}.
Using no factor (A0) degrades the model to the GT-Aug baseline.
Even with one factor, COMAug (A1-A4) can already result in noticeable performance gains.
The best performing single factor A2 indicates that distance is the main contributor to object difficulty in LiDAR object detection.
Improvements from A5 to A8 increase steadily as more factors are taken into account.
Final results are of improvements 0.98\% and 0.95\% on AP$_{L1}$ and AP$_{L2}$, respectively, when four factors are utilized.

\begin{figure}[ht!]
  \centering
  {\includegraphics[width=1.0\linewidth]{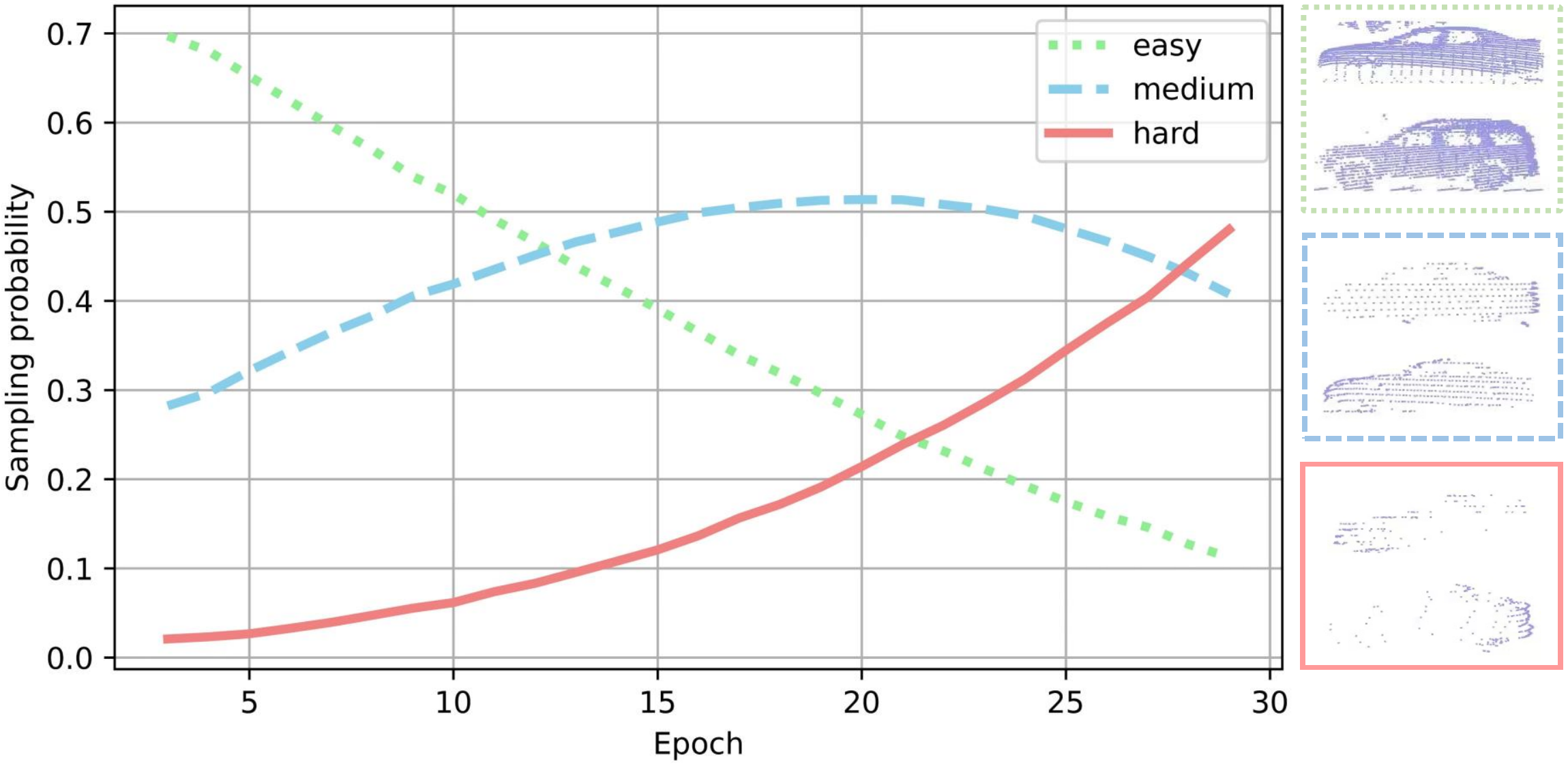}}
  \caption{Visualization of sampling probability.
  } \label{fig:probability3}
  \vspace{-5mm}
\end{figure}

\noindent
\textbf{Visualization of sampling probability.}
Based on the difficulties predicted in the last epoch, we categorize the groups into easy/medium-hard/hard parts which have top 1/3, mid 1/3 and bottom 1/3 of difficulty scores.
Consequently, we visualize the mean sampling probabilities for each part as shown in \cref{fig:probability3}.
Note that we ignore the first epoch where all probabilities are initialized to be the same.
The trends of these curves are shown as expected:
easy groups are overwhelmingly sampled in early stages while rarely sampled in the end.
The sampling for hard groups is in an opposite pattern.
For medium groups, the sampling probabilities increase first and then decrease, and peak in the middle of training.

\begin{table}[ht!]
  \centering
\renewcommand{\arraystretch}{1.05}
\renewcommand{\tabcolsep}{3pt}
  \small
  \begin{tabular}{c|cccc|ll}
    \hline
    & Occup. & Dist.  & Angle & Size  & \multicolumn{1}{c}{3D AP$_{L1}$ (\%)} & \multicolumn{1}{c}{3D AP$_{L2}$ (\%)} \\
    \hline
    A0 & & & & & 71.05 & 62.66\\
    \hline
    A1 & {\textbf \checkmark} &&&  & 71.36 \improvea{0.31} & 62.93 \improvea{0.33}\\
    A2 && \checkmark &&  & 71.51 \improvea{0.46} & 63.11 \improvea{0.45} \\
    A3 &&& \checkmark &  & 71.33 \improvea{0.28} & 62.92 \improvea{0.32} \\
    A4 &&&& \checkmark   & 71.38 \improvea{0.33} & 62.98 \improvea{0.38} \\
    A5 & \checkmark & \checkmark & &  & 71.68 \improvea{0.63} & 63.24 \improvea{0.31}\\
    A6 & \checkmark & \checkmark & \checkmark &  & 71.86 \improvea{0.81} & 63.41 \improvea{0.75} \\
    A7 & \checkmark & \checkmark && \checkmark & 71.89 \improvea{0.84} & 63.44 \improvea{0.78}\\
    A8 & \checkmark & \checkmark & \checkmark & \checkmark  & 72.03 \improvea{0.98} & 63.61 \improvea{0.95}\\        
  \bottomrule         
  \end{tabular}
  \caption{Performances with different clustering factors.
    Occup. and Dist. stand for  occupancy ratio and distance, respectively.
  }\label{tab:clusterfactor} 
\end{table}

\section{Conclusions and Limitations}\label{sec:conclusion}
In this work, we explore the potentials of curriculum learning in LiDAR-based object detection by proposing a novel curricular object manipulation (COM) framework.
We focus on loss design in the detector and sampling strategy in GT-Aug, and correspondingly introduce the COMLoss and COMAug modules.
Extensive experiments and ablation studies on large scale benchmarks verified the efficacy of our method.

However, there is still room for future research on applying curriculum learning to 3D object detection.
To inspire future work, we outline a few limitations based on our understanding:
(1) our difficulty criterion depends sorely on the classification loss while the regression difficulty is ignored for the sake of efficiency.
Moreover, loss-based criterion is naturally affected by training noise and hysteresis effect, and therefore may not precisely reveal difficulties.
Designing more accurate criterion is a promising direction.
(2) we group objects by four empirically validated heuristics.
More efforts are deserved for more appropriate grouping strategies as well.
(3) we limit the work in the LiDAR-based object detector with GT-Aug.
This work can be extended to other LiDAR-related tasks such as multi-modality 3D object detection and point cloud segmentation.



\clearpage


\rule[0.25\baselineskip]{0.1\textwidth}{1pt}
Supplemental File
\rule[0.25\baselineskip]{0.1\textwidth}{1pt}

\maketitle
\section{Supplementary}
  In the supplemental file, we provide more details of our work to supply our main paper, including
  \begin{itemize} 
 \item \textbf{Details of our object clustering} about the occupancy ratio and the clustering rules in our experiments,
 \item \textbf{More visualization results} of our clustering rules and different clustering groups,
  \item \textbf{Deeper analysis of COMAug} about the influence of the hyper parameters and the effectiveness of anti-curriculum learning.
 
 \end{itemize}

\subsection{Details of our object clustering}
 
\noindent\textbf{Explanation of occupancy ratio.}
In our main paper,
we have clarified that the occupancy ratio $f_{o}$ is the ratio of non-empty 3D voxels inside the bounding box of an object.
Considering the shape differences of the point cloud objects,
we employ different voxel division strategies for vehicles and pedestrians
in our experiments.
\begin{figure}[htb!]
  \centering
  {\includegraphics[width=1\linewidth]{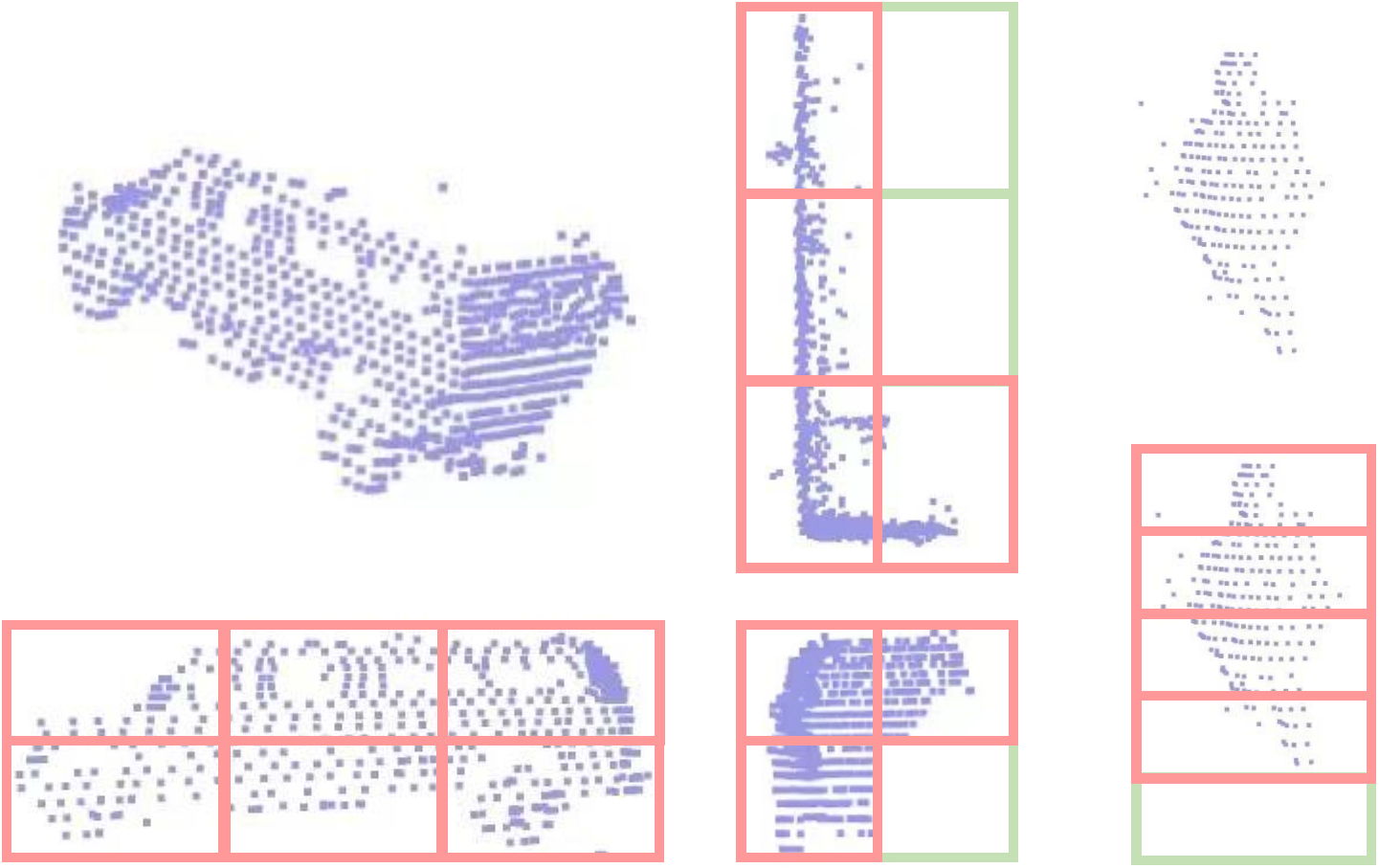}}
  \caption{Voxel divisions for the Vehicle and Pedestrian.
  Red boxes are non-empty boxes and the green ones are empty.
  Occupancy ratio denotes the ratio of non-empty boxes.}
  \label{fig:ratio}
\end{figure}

We illustrate our voxel division strategies in \figref{fig:ratio}.
For the vehicle, 
we divide the bounding box into $3\times2\times2$ voxels along the length, width, and height. 
For Pedestrian we divide the box into 5 voxels by height.

\noindent\textbf{The clustering rules in our experiments.}
We cluster objects in the ground-truth database into groups by the distance $f_{d}$, size $f_{s}$, relative angle $f_{a}$,  and occupancy ratio $f_{o}$.
In our experiments, 

\noindent(1) the distances of objects are divided into $[0m,30m)$,
$[30m,50m)$, and $[50m,+\infty)$.

\noindent(2) Following \cite{fan2022embracing}, the sizes of the vehicles are divided into  
$[0m,4m)$,
$[4m,8m)$, and $[8m,+\infty)$.

\noindent(3) For the relative angle $f_{a}$,
we divide $f_{a}(mod\frac{\pi}{2})$ of the vehicles into $[0,\frac{\pi}{6})$,
$[\frac{\pi}{6},\frac{\pi}{3})$, and $[\frac{\pi}{3},\frac{\pi}{2}]$.

\noindent (4) The occupancy ratios of the objects range from 0 to 1.
We divide them into $[0,0.2)$,
$[0.2,0.4)$, $[0.4,0.6)$, $[0.6,0.8)$, and $[0.8,1]$.

Importantly, in the Tab. 1 of our main paper, we use the distance and occupancy ratio to cluster pedestrians.
In this way, the pedestrians are clustered into 15 groups.
In the Tab. 2 of our main paper, all of the factors are utilized and the vehicles are clustered into 135 groups.
In the Tab. 3 of our main paper, 
different combinations of (1), (2), (3), and (4) are implemented for clustering the vehicle class.

\subsection{A deeper analysis of COMAug}

\begin{figure}[htb!]
  \centering
  {\includegraphics[width=1\linewidth]{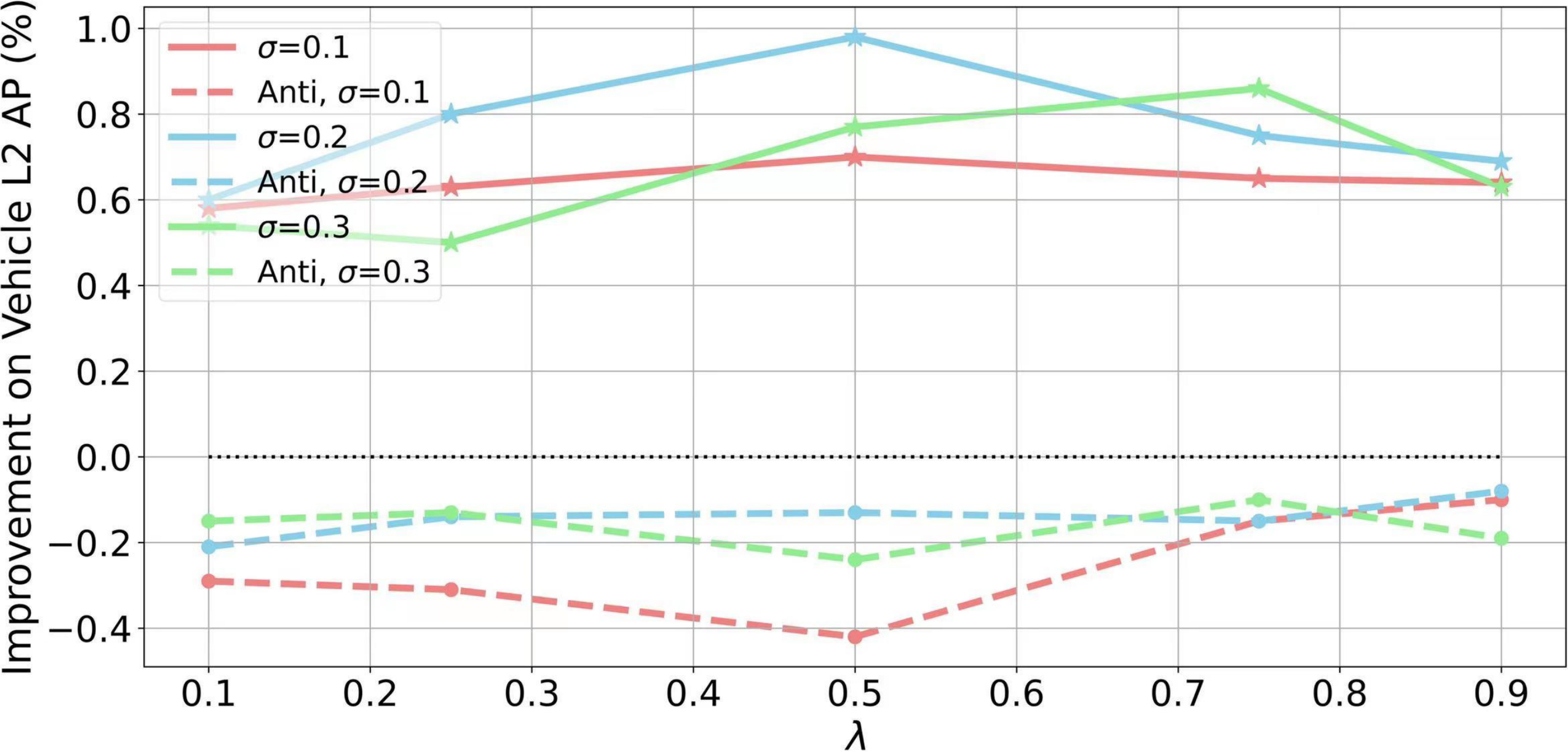}}
  \caption{The performance improvement with different hyper parameters settings in COMAug.
  Solid lines represent curriculum learning, while dashed lines represent anti-curriculum learning.}
  \label{fig:anti}
\end{figure}

\noindent\textbf{Effects of the parameter $\lambda$ and $\sigma$.}
The $\lambda$ in our COMAug controls the pacing speed, 
which determines how early the probabilities of sampling semi-hard/hard objects are raised.
The pacing speed is faster with a larger $\lambda$.
As is shown in \figref{fig:anti},
when $\sigma=0.1, 0.2$, setting $\lambda$ to 0.5 can lead to better performances. 
When $\sigma=0.3$, assigning $\lambda$ with 0.75 is optimal.
Overall, the $\lambda$ should be adjusted to a moderate value,
where the pacing speed is neither too fast nor too slow.
This conclusion is similar to that in other curriculum learning works~\cite{kong2021adaptive, huang2020curricularface}.

The parameter $\sigma$ control the diversity of the sampling.
When $\sigma \xrightarrow{} 0$, only one group whose score $s_i$ equals $\mu_t$ will be sampled at each training epoch $t$.
When $\sigma \xrightarrow{} \infty$, every group will have an equal probability of being sampled.
In \figref{fig:anti},
we set $\sigma$ to 0.1, 0.2, and 0.3.
The experiments show that $\sigma = 0.2$ can typically achieve better performances.

\noindent\textbf{Effects of anti-curriculum learning.}
In the introduction of our main paper,
we claim that selecting too many hard
examples at early stages may overwhelm the training, 
while selecting too many easy samples at the later stages may
slow the model convergence.
Here we employ anti-curriculum COMAug to validate this point.

In contrast to curriculum learning, anti-curriculum learning emphasizes training samples from difficult to easy.
Even if counter-intuitive, \cite{kocmi2017curriculum, zhang2018empirical, zhang2019curriculum} have shown the effectiveness of anti-curriculum in certain scenarios.
However, most works in curriculum learning demonstrate the superiority of curriculum than anti-curriculum or random ordering.

We conduct experiments to validate whether anti-curricular can be effective for our task.
Specifically,
we reverse the sorted group scores $\{s_i\}_{1}^{G}$,
so that our COMAug selects increasingly easy objects for augmentation as training proceeds.
The settings of $\lambda$ and $\sigma$ are kept the same as the curricular experiments in \figref{fig:anti}.
Through the experiment results,
we find that anti-curriculum is inferior to standard training.
Notice that when $\lambda = 0.1$ and $\sigma=0.1$, 
many difficult objects are sampled during the early training stage.
In this case, anti-curriculum brings a 0.27\% performance drop.
This phenomenon demonstrates that the hard-sample-first strategy can hinder convergence.
Moreover, 
However, the performances are still worse than the standard training.
It shows that selecting too many easy objects during the later stage does not benefit the training.

\noindent\textbf{The minor improvement lead by COM on Pedestrian class.}
The improvements of COMAug+COMLoss over COMAug are not as significant for class pedestrian as those for class vehicle.
The phenomenon could be caused by data distribution as shown in Tab.~\ref{tab:1}.
For vehicle, each frame contains 30.2 source objects and 1.8 augmented objects on average.
In contrast, numbers for pedestrian are 14.1 and 5.9.
Recall that COMLoss essentially improves model performances by treating objects differently based on their difficulties.
COMAug, on the other hand, selects objects of similar difficulty levels (suitable for current training) for augmentation.
That decreases the variety of object difficulties and weakens the effects of COMLoss.
As a result, the improvements for class pedestrian are less evident due to its high ratio of augmented objects (29.5\%), while  class vehicle is less affected because of its low ratio (5.6\%).
Besides, we found that the phenomenon does not occur for GT-Aug.
The reason could be that GT-Aug selects objects at random, which results in diverse object difficulties.

\begin{table}[htb!]
\centering
\footnotesize
\renewcommand{\arraystretch}{1.0}
\renewcommand{\tabcolsep}{1.5mm}
\caption{Average number of objects per frame in Waymo Dataset.} \label{tab:1}
\begin{tabular}{l|cc|c}
		\hline
	    &  Source &  Augmented &  Ratio of   augmented  object \\
		\hline
	    Vehicle & 30.2 & 1.8 & 5.6\% \\
	    Pedestrian & 14.1 & 5.9 & 29.5\%   \\
		\hline
\end{tabular}
\end{table}

\subsection{More visualization results}

\begin{figure}[t!]
  \centering
  {\includegraphics[width=1\linewidth]{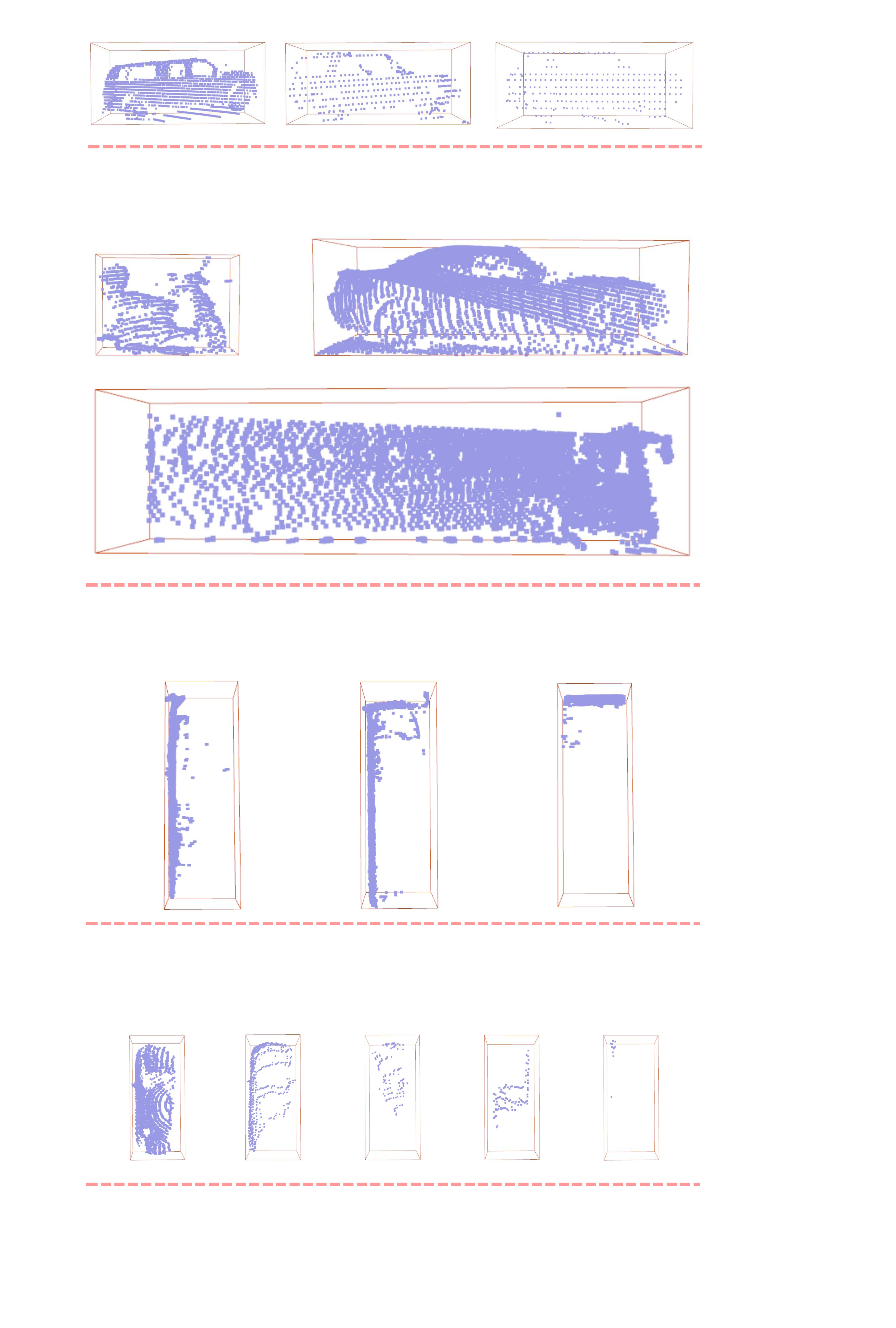}}
  \put(-150, -10){Occupancy Ratio}
  \put(-145, 90){Relative Angle}
  \put(-127, 220){Size}
  \put(-135, 390){Distance}
  \vspace{5pt}
  \caption{Visualization of the vehicles with different distances, sizes, relative angles, and occupancy ratios.}
  \label{fig:vis0}
\end{figure}

In this work,
we cluster the vehicles according to four factors: distance, size, relative angle, and occupancy ratio.
To demonstrate how the four factors affect the distribution of the vehicle point clouds, 
we present a few examples in \figref{fig:vis0}. 
We can observe the following four phenomena:
(1) The objects' point clouds become sparser as the distance increases.
(2) The sizes and shapes of the vehicles can vary greatly.
(3) Our relative angle can reveal the faces of the vehicle being observed.
(4) Objects with low occupancy ratios can be difficult to recognize.

\begin{figure*}[htb!]
  \centering
  {\includegraphics[width=0.85\linewidth]{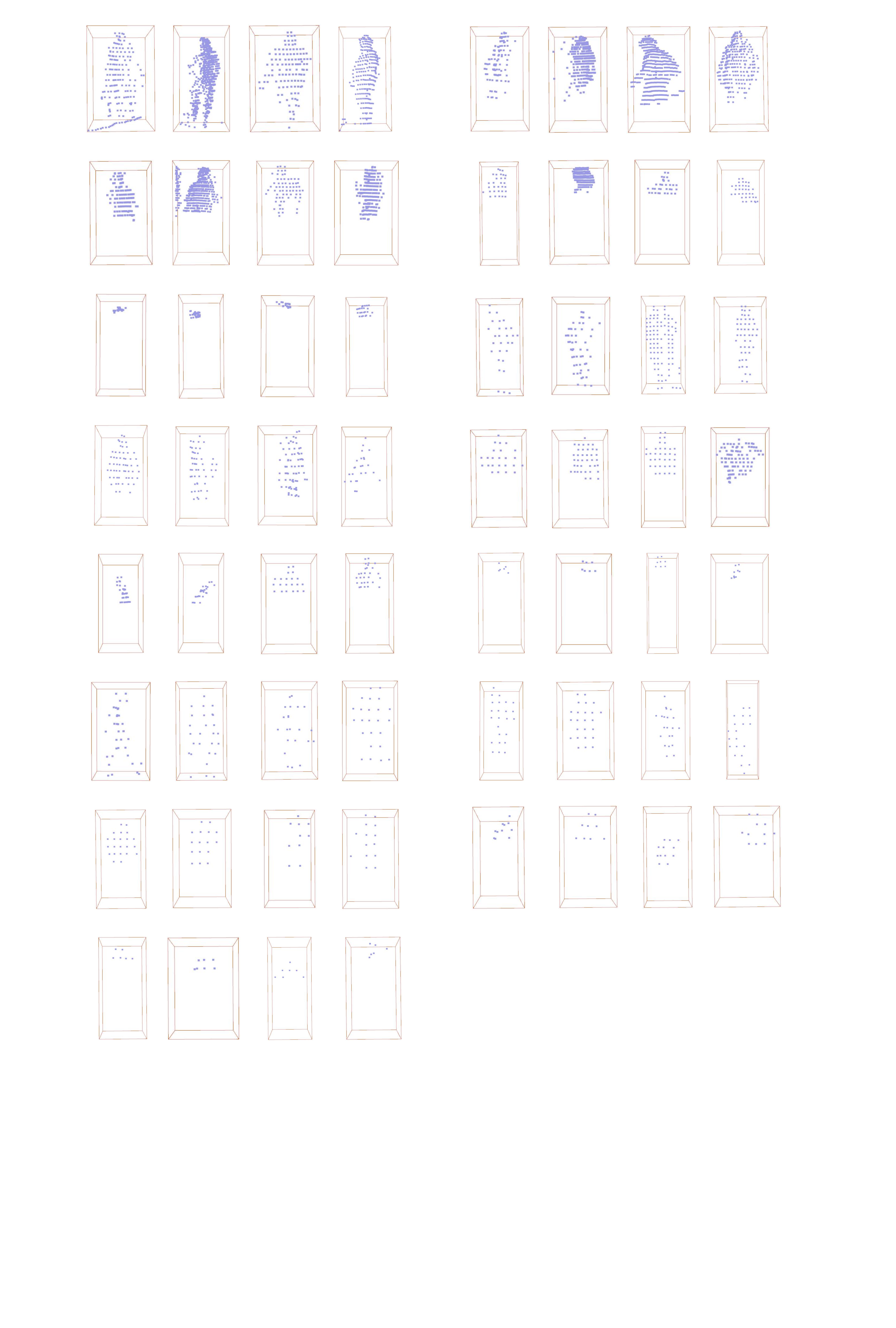}}
  \caption{Visualization of the 15 pedestrian groups.}
  
  \label{fig:vis1}
\end{figure*}

Further, we cluster the pedestrians into 15 groups according to the distance and occupancy ratio in the Tab. 1 of our main paper.
In \figref{fig:vis1} we provide the visualization examples of the pedestrian groups.
The sizes of the pedestrians are typically similar,
but their point clouds vary widely in sparsity and shape integrity.


\clearpage

{\small
\bibliographystyle{ieee_fullname}
\bibliography{egbib}
}

\end{document}